\def\year{2022}\relax
\documentclass[letterpaper]{article} 
\usepackage{aaai22}  
\usepackage{times}  
\usepackage{helvet}  
\usepackage{courier}  
\usepackage[hyphens]{url}  
\usepackage{graphicx} 
\usepackage{natbib}  
\usepackage{caption} 
\DeclareCaptionStyle{ruled}{labelfont=normalfont,labelsep=colon,strut=off} 
\usepackage{algorithm}
\usepackage{algorithmic}
\usepackage[table]{xcolor}
\usepackage{etoolbox}
\usepackage{pgf}
\usepackage{hhline}

\definecolor{high}{HTML}{228b22}  
\definecolor{low}{HTML}{ffffff}  
\newcommand*{\opacity}{90}
\newcommand*{\minval}{0.98}
\newcommand*{\maxval}{8.26}
\newcommand{\gradient}[1]{
    \ifdimcomp{#1pt}{>}{\maxval pt}{#1}{
        \ifdimcomp{#1pt}{<}{\minval pt}{#1}{
            \pgfmathparse{int(round(100*(#1/(\maxval-\minval))-(\minval*(100/(\maxval-\minval)))))}
            \xdef\tempa{\pgfmathresult}
            \cellcolor{high!\tempa!low!\opacity} #1
    }}
}

\usepackage{newfloat}
\usepackage{listings}
\floatstyle{ruled}
\newfloat{listing}{tb}{lst}{}
\floatname{listing}{Listing}

\usepackage{bm}
\usepackage{amsmath}
\DeclareMathOperator*{\argmax}{arg\,max}

\title{Learning and Understanding a Disentangled Feature Representation for Hidden Parameters in Reinforcement Learning}
\author{
    Christopher Reale, Rebecca Russell
}
\affiliations{
    Draper\\

    555 Technology Square\\
    Cambridge, MA, 02139\\
    creale@draper.com, rrussell@draper.com
}

\begin{document}

\maketitle

\begin{abstract}
    Hidden parameters are latent variables in reinforcement learning (RL) environments that are constant over the course of a trajectory. Understanding what, if any, hidden parameters affect a particular environment can aid both the development and appropriate usage of RL systems. We present an unsupervised method to map RL trajectories into a feature space where distance represents the relative difference in system behavior due to hidden parameters. Our approach disentangles the effects of hidden parameters by leveraging a recurrent neural network (RNN) world model as used in model-based RL. First, we alter the standard world model training algorithm to isolate the hidden parameter information in the world model memory. Then, we use a metric learning approach to map the RNN memory into a space with a distance metric approximating a bisimulation metric with respect to the hidden parameters. The resulting disentangled feature space can be used to meaningfully relate trajectories to each other and analyze the hidden parameter. We demonstrate our approach on four hidden parameters across three RL environments. Finally we present two methods to help identify and understand the effects of hidden parameters on systems.
\end{abstract}

\section{Introduction}
\label{intro}

\begin{figure}[ht!]
\begin{center}
\centerline{\includegraphics[width=0.7\columnwidth]{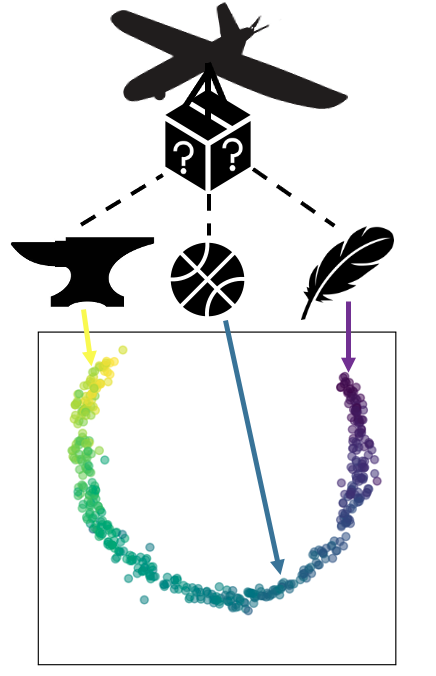}}
\caption{Our method maps trajectories to a feature space that represents the hidden parameters of the system. For example, we can map the trajectory of a UAV with an unknown payload mass (top) into a disentangled feature space (bottom).}
\label{fig-intro}
\end{center}
\end{figure}

Many robotic skills require the ability to operate effectively in similar but slightly-varying environments. For example, when controlling a robot to move through an environment, it must be able to operate on surfaces with different coefficients of friction. Like friction, many conditions that affect system behavior are fixed over a long period of time, perhaps even the entire course of a trajectory, but may vary from one trajectory to the next. Termed \emph{hidden parameters} \citep{doshi2016hidden}, these conditions can affect the behavior and performance of an RL system. In many real-world applications, it may not be clear how many hidden parameters exist or how they affect the system dynamics. In this work, we present an unsupervised method to learn feature representations of trajectories by their disentangled hidden parameter values. This feature space can then be used in downstream classification or analysis tasks to aid in interpreting the hidden parameters.

Our approach leverages world models (also termed ``dynamics models'') used in deep model-based reinforcement learning (MBRL) systems, though it is broadly applicable dynamical systems not based on MBRL.  In recent years, deep MBRL has emerged as a data-efficient learning approach to a wide variety of tasks \citep{nagabandi2018neural}. In deep MBRL, a world model represented by a neural network is trained to predict the next state given the current state and an action. The world model is then used with model-predictive control or model-free reinforcement learning to select actions given a reward function. Partially-observable state representations are typically handled by including a recurrent layer or layers in the world model neural network architecture, allowing the model to incorporate a memory of state observations in its predictions. The features learned in this world model memory should include information on any hidden parameters mixed in with other transient and irrelevant information.

In our previous work \cite{reale2022workshop}, we presented two innovations which enable the learning of a disentangled feature space to represent hidden parameters. Our first innovation is to modify the world model training algorithm to constrain part of the world model's internal recurrent representation (memory) to be \textit{time-invariant}: the model's predictions should not change when the values of the time-invariant features are replaced with their corresponding values at another time-step along the same trajectory. This constraint essentially prohibits the world model from storing transient information in the time-invariant features, leaving only information pertaining to the hidden parameters.

Our second innovation is a metric learning approach to map the time-invariant features into a space with a meaningful metric that we term a \textit{latent bisimulation metric}. While the time-invariant features contain only hidden parameter information, it is not possible to directly relate two sets of time-invariant features. For example, two hidden parameter representations may be far away in time-invariant feature space but have a similar effect on the predictions produced by the world model and thus represent a similar pair of hidden parameter values. The only way to relate two of time-invariant representations in a semantically meaningful way is through the usage of the world model. To fix this, we learn a mapping from the time-invariant feature space into a metric space with distance proportional to the difference in system behaviors. This is akin to learning a bisimulation metric \citep{ferns2011bisimulation} on the latent state (\emph{i.e.}, the hidden parameters) as opposed to the observable state.

In this paper, we summarize our previous work and further validate our approach by applying it to a more complex UAV application. Additionally, we present two methods to help explain the effects of the unknown hidden parameter on a system to a user.

\section{Related Work}
\subsection{Hidden Parameters}

While the control of systems in partially-observable environments has been a focus of research dating back decades \citep{kaelbling1998planning}, only recently did \citet{doshi2016hidden} introduce the idea of studying constant unobservable variables, termed ``hidden parameters.'' They contend that modeling a hidden parameter system, as opposed to an arbitrary partially-observable system, simplifies the procedure for inferring the system dynamics. They also present an approach for modeling the dynamics of a hidden parameter system from data. \citet{killian2017robust} improved on their approach by incorporating a Gaussian Process latent variable model to improve uncertainty quantification and transfer. \citet{yao2018direct} further improve transfer-ability by embedding the hidden parameter values as an input to a policy. While many environments of interest may contain multiple hidden parameters of different types (e.g., continuous vs. discrete) as well as additional dynamic latent variables, we focus on the simple case of environments with a single discrete valued hidden parameter and no other latent variables. We leave the analysis of more complex hidden parameter spaces to future work.

While there is relatively little work specifically analyzing hidden parameters, there is a long and vast history of work related to the more general scenario of Partially Observable Markov Decision Processes (POMDPs) such as \citet{spaan2012partially}; \cite{cassandra1998survey}; \citet{oliehoek2012influence}; and \citet{carr2021task}.

\subsection{Model-Based Reinforcement Learning}

Model-Based Reinforcement learning is one of the main learning-based approaches to controlling systems. Initially, most model based reinforcement learning systems used simple world models such as Gaussian Processes \citep{boedecker2014approximate, deisenroth2011pilco, ko2009gp},  time-varying linear models \citep{levine2014learning, lioutikov2014sample, yip2014model}, and mixture of Gaussians \citep{khansari2011learning}. While data efficient, these approaches have trouble modeling high-dimensional systems with non-linear dynamics. Recently, it has been shown that more complex deep neural networks can be used effectively as well \citep{akkaya2019solving, nagabandi2018neural, nagabandi2020deep}.

Since hidden parameter systems are partially observable, a world model cannot accurately predict a state from the preceding state and action alone. Instead, world models must infer information about the unobserved latent variable(s) through interactions with the environment over time. Recurrent neural networks (RNN) have been used as world models to enable this functionality \citep{schmidhuber1990line, schmidhuber1990making, schmidhuber2015learning, ha2018recurrent}.

\subsection{Bisimulation Metrics}
Bisimulation metrics \citep{ferns2011bisimulation, ferns2014bisimulation} measure the behavioral similarity of two states of a Markov decision process (MDP). Much research has focused on methods to compute bisimulation metrics \citep{taylor2008bounding,castro2020scalable}. Similar to our approach, \citet{zhang2020learning} proposed using bisimulation metrics to learn a feature representation in a reinforcement learning setting. Our approach differs by learning a bisimulation metric on the latent portion of the state rather than the observable portion. Additionally, we use the learned representation to better understand hidden parameters, whereas \citet{zhang2020learning} sought to improve control by filtering out irrelevant information.

\section{Background}

\subsection{Model-Based Reinforcement Learning}

Consider an agent operating in an environment. Let $\bm{s}_t \in \mathcal{S}$ be the observable environmental state and $\bm{a}_t \in \mathcal{A}$ be the action taken by the agent at time $t$. At every time step, the system observes state $\bm{s}_t$, takes action $\bm{a}_t$, and obtains the reward $r(\bm{s}_t, \bm{a}_t)$. The system then takes the observable state $\bm{s}_{t+1}$ according to the system transfer function $f : \mathcal{S} \times \mathcal{A} \rightarrow \mathcal{S}$. The goal in reinforcement learning is to select an action sequence $\bm{A}_t^{(H)} = (\bm{a}_t, \bm{a}_{t+1}, \ldots, \bm{a}_{t+H-1})$ to maximize rewards over a horizon length $H$, 

\begin{equation}
\begin{aligned}
\bm{A}_t^{(H)} = \argmax_{\bm{A}_t^{(H)}} \sum_{t'=t}^{t+H-1} &r(\bm{s}_{t'}, \bm{a}_{t'}) \\
&\bm{s}_{t'+1} = f(\bm{s}_{t'},\bm{a}_{t'}).
\end{aligned}
\end{equation}

In MBRL, the transfer function $f$ is explicitly modeled as $\hat{f}_\theta$ with parameters $\theta$. Regardless of the model chosen, its parameters are learned from example trajectories and, at run-time, it is used select optimal actions as follows:
\begin{equation}
\begin{aligned}
\bm{A}_t^{(H)} = \argmax_{\bm{A}_t^{(H)}} \sum_{t'=t}^{t+H-1} &r(\bm{\hat{s}}_{t'}, \bm{a}_{t'}) \\
&\bm{\hat{s}}_{t'+1} = \hat{f}_\theta(\bm{\hat{s}}_{t'},\bm{a}_{t'}).
\end{aligned}
\end{equation}

\subsection{World Models}
\label{world_models}

In deep MBRL, the world model $\hat{f}_\theta$ is a deep neural network with network weights $\theta$. In our case, we use a recurrent neural network architecture to be able to operate in a partially-observable environment. We represent the internal state of the RNN at time $t$ as $\bm{h}_t$. The model parameters $\theta$ are learned by stochastic gradient descent to minimize the prediction error over a dataset $\mathcal{D}$ of trajectories of lengths $T_d$ collected with a random policy:
\begin{equation}
\min_{\theta} \frac{1}{|\mathcal{D}|} \sum_{d=0}^{|\mathcal{D}|-1}\frac{1}{T_d-1}\sum_{t=1}^{T_d-1} \|\bm{s}_{t} - \bm{\hat{s}}_{t}\|_{1} \\
\end{equation}
where
\begin{equation*}
\begin{aligned}
\bm{\hat{s}}_{t+1} &= \hat{f}_\theta(\bm{s}_{t},\bm{a}_{t}, \bm{h}_{t}) \\
\bm{h}_{t+1} &= g_\theta(\bm{s}_{t},\bm{a}_{t}, \bm{h}_{t}) \\
\bm{h}_{0} &= \bm{0}
\end{aligned}
\end{equation*}

\section{Our Approach}

Our goal is to learn a feature space that represents trajectories by the hidden parameters of the environment in which they occurred. Our approach consists of two main steps. First, we present a new deep reinforcement learning world model training algorithm to isolate time-invariant information related to the hidden parameters in the RNN memory. Then, we show how to learn a mapping from the RNN memory to a feature space where the distance between points approximates a bisimulation metric.

\subsection{Time-Invariant RNN Memory}
\label{section-invariant}

\begin{figure*}[ht]
\begin{center}
\centerline{\includegraphics[trim={2cm, 6cm, 2cm, 6cm}, clip, width=\textwidth]{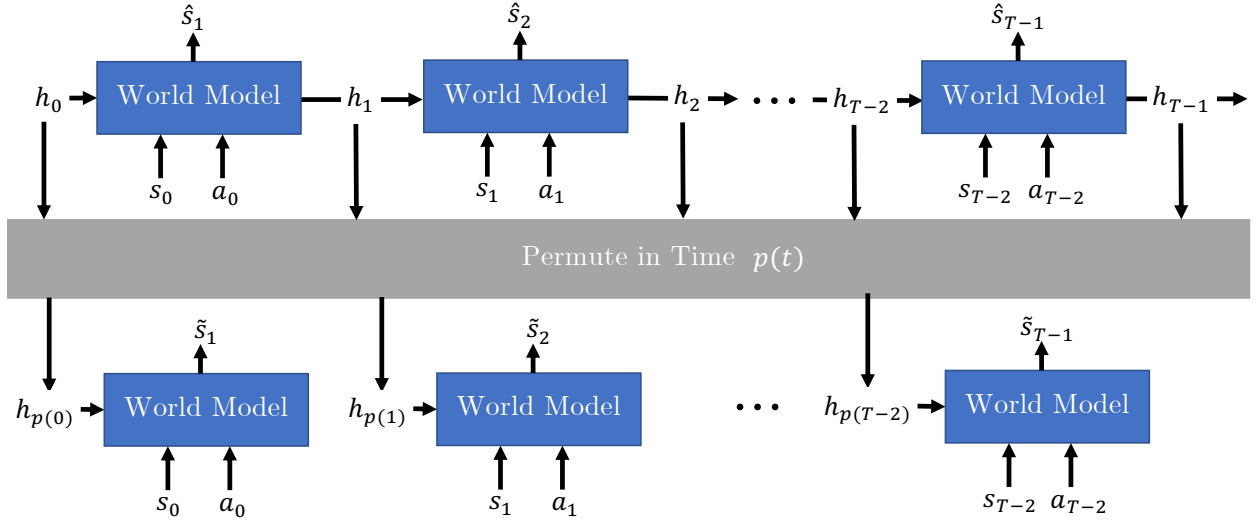}}
\caption{Visualization of time-invariant training procedure shuffling the world model RNN memory in time over a trajectory.}
\label{fig-time-invariant}
\end{center}
\end{figure*}

\begin{figure*}[ht]
\begin{center}
\centerline{\includegraphics[trim={0cm, 7cm, 0cm, 6cm}, clip, width=\textwidth]{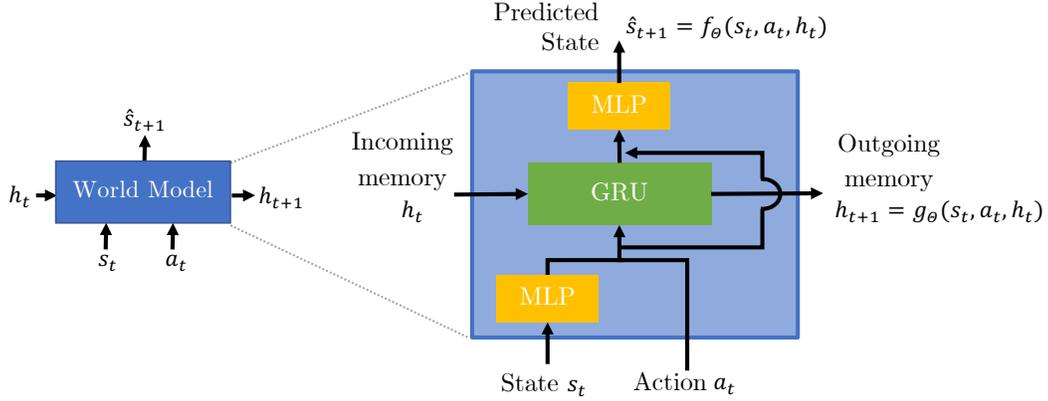}}
\caption{Recurrent neural network world model architecture.}
\label{fig-world-model}
\end{center}
\end{figure*}

When operating in an environment with hidden parameters, the world model cannot accurately predict future states without information about the parameters. An RNN world model acquires this information by interacting with the environment and encoding its trajectories in memory ($\bm{h}_t$) to enable better predictions in the future. While the memory is the only source of hidden parameter information the world model has access to, it also may contain other transient information that helps the world model make predictions (e.g., past states and actions). In this section, we present an approach to encode information related to hidden parameters only.

The key insight to our approach is that, since hidden parameters are constant over the course of a trajectory, features computed at a given time step that only contain information about hidden parameters should be usable at all times along the trajectory. Inversely, if the features contain additional information related to transient factors, they will not be usable at all times along the trajectory. Thus, we seek to train the RNN to encode features in its memory that are applicable to the rest of the trajectory. We term this \textit{time-invariant memory}.

We achieve this by adding an additional step to the training procedure presented earlier. As before, we generate a dataset of trajectories and use stochastic gradient descent to minimize prediction error. However, in addition to making predictions ${\hat{s}}_{t+1}$ of the next states using the temporally-appropriate RNN memory values $\bm{h}_{t}$, we also make predictions ${\tilde{s}}_{t}$ using RNN memory values from other time steps as shown in Figure \ref{fig-time-invariant}. More formally, we define $\bm{h}_{p(t)}$ where $p : \{0,1,\ldots, T_d\} \rightarrow \{0,1,\ldots, T_d\}$ is a random bijection that serves to permute the hidden memory values in time. Thus, we minimize prediction error by:

\begin{equation}
\min_{\theta} \frac{1}{|\mathcal{D}|} \sum_{d=0}^{|\mathcal{D}|-1}\frac{1}{T_d-1}\sum_{t=1}^{T_d-1} \|\bm{s}_{t} - \bm{\hat{s}}_{t}\|_{1} + \|\bm{s}_{t} - \bm{\tilde{s}}_{t}\|_{1} 
\end{equation}

where

\begin{equation*}
\begin{aligned}
\bm{\hat{s}}_{t+1} &= \hat{f}_\theta(\bm{s}_{t},\bm{a}_{t}, \bm{h}_{t}) \\
\bm{\tilde{s}}_{t+1} &= \hat{f}_\theta(\bm{s}_{t},\bm{a}_{t}, \bm{h}_{p(t)}) \\
\bm{h}_{t+1} &= g_\theta(\bm{s}_{t},\bm{a}_{t}, \bm{h}_{t}) \\
\bm{h}_{0} &= \bm{0}
\end{aligned}
\end{equation*}
This training procedure encourages the world model to rapidly populate its memory from a sequence of observations at any time with information that is useful at all times in a trajectory.

\subsection{Latent Bisimulation Metric}
\label{section-metric}
Ultimately, our goal is to learn a feature representation to be able to directly compare two trajectories in terms of the hidden parameter values of their corresponding environments. While our time-invariant RNN memory contains information pertaining only to the hidden parameters, the RNN memory from different time steps and/or trajectories may not be directly comparable. For example, two points in the RNN memory feature space may have very different values, yet have a similar effect on the world model predictions and represent the same hidden parameter. In this section we present a metric learning approach to map from the time-invariant RNN-memory space with distance proportional to the differences in system behavior. Since our feature space encodes differences in system behavior based on latent variables (hidden parameters) only, we call our learned metric a \textit{latent bisimulation metric}.

We learn the embedding in two steps. First, we compute pairwise distances between RNN memory features as the average prediction difference over a representative set of state-action pairs. Then, we train a neural network to map the RNN memory features into a space that enforces those distances.

In order to compute the pairwise distances, we first create a set of state-action pairs, $\mathcal{P} = \{(\bm{s}_0, \bm{a}_0), (\bm{s}_1, \bm{a}_1), \ldots , (\bm{s}_{P-1}, \bm{a}_{P-1})\}$ by sampling them from a large set of trajectories. We then create the input to our training dataset: a representative set of RNN memory values $\mathcal{H} = \{\bm{h}_0, \bm{h}_1, \ldots, \bm{h}_{H-1}\}$, again by sampling from a large set of trajectories. Finally, we compute the distances $d(\bm{h}_i, \bm{h}_j)$ between points as follows:

\begin{equation}
d(\bm{h}_i, \bm{h}_j) = \frac{1}{P}\sum_{p=0}^{P-1} \|\hat{f}_\theta(\bm{s}_{p},\bm{a}_{p}, \bm{h}_{i})-\hat{f}_\theta(\bm{s}_{p},\bm{a}_{p}, \bm{h}_{j})\|_1
\end{equation}

The distance measures the average (over a representative set of pairs of states and actions) difference in predictions made by the world model when using the two memory values. Once we have computed the distances, we learn an embedding function $e_\phi(\bm{h})$, modeled as a neural network with parameters $\phi$, by solving the following optimization problem with stochastic gradient descent:

\begin{equation}
\min_\phi \sum_{i=0}^{H-1}\sum_{j=i}^{H-1} \left|\|e_\phi(\bm{h}_i)-e_\phi(\bm{h}_j)\|_1 - d(\bm{h}_i,\bm{h}_j)\right|
\end{equation}

\section{Experiments}

\subsection{Use Cases}

\subsubsection{Modified Mountain Car}
\begin{figure}[ht]
\begin{center}
\centerline{\includegraphics[width=0.8\columnwidth]{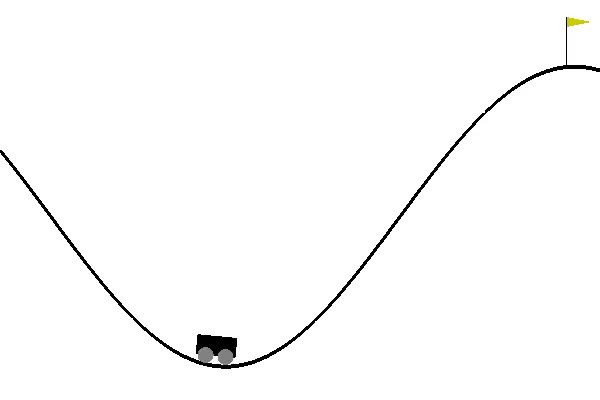}}
\caption{Visualization of the mountain car environment.}
\label{fig-mountaincar}
\end{center}
\end{figure}

In the mountain car \citep{brockman2016gym} RL problem, shown in Figure \ref{fig-mountaincar}, the goal is for the car to reach the top of the mountain. The state consists of two variables: horizontal position and horizontal velocity. The action space consists of three possible actions: accelerate to the right, coast, and accelerate to the left. We modified the mountain car environment to include variable gravity as a hidden parameter. For each trajectory, the gravity strength is randomly chosen from a set of three values: 75\%, 100\%, and 125\% of the original gravity.

\subsubsection{Pusher Robot}
\begin{figure}[ht]
\begin{center}
\centerline{\includegraphics[width=0.8\columnwidth]{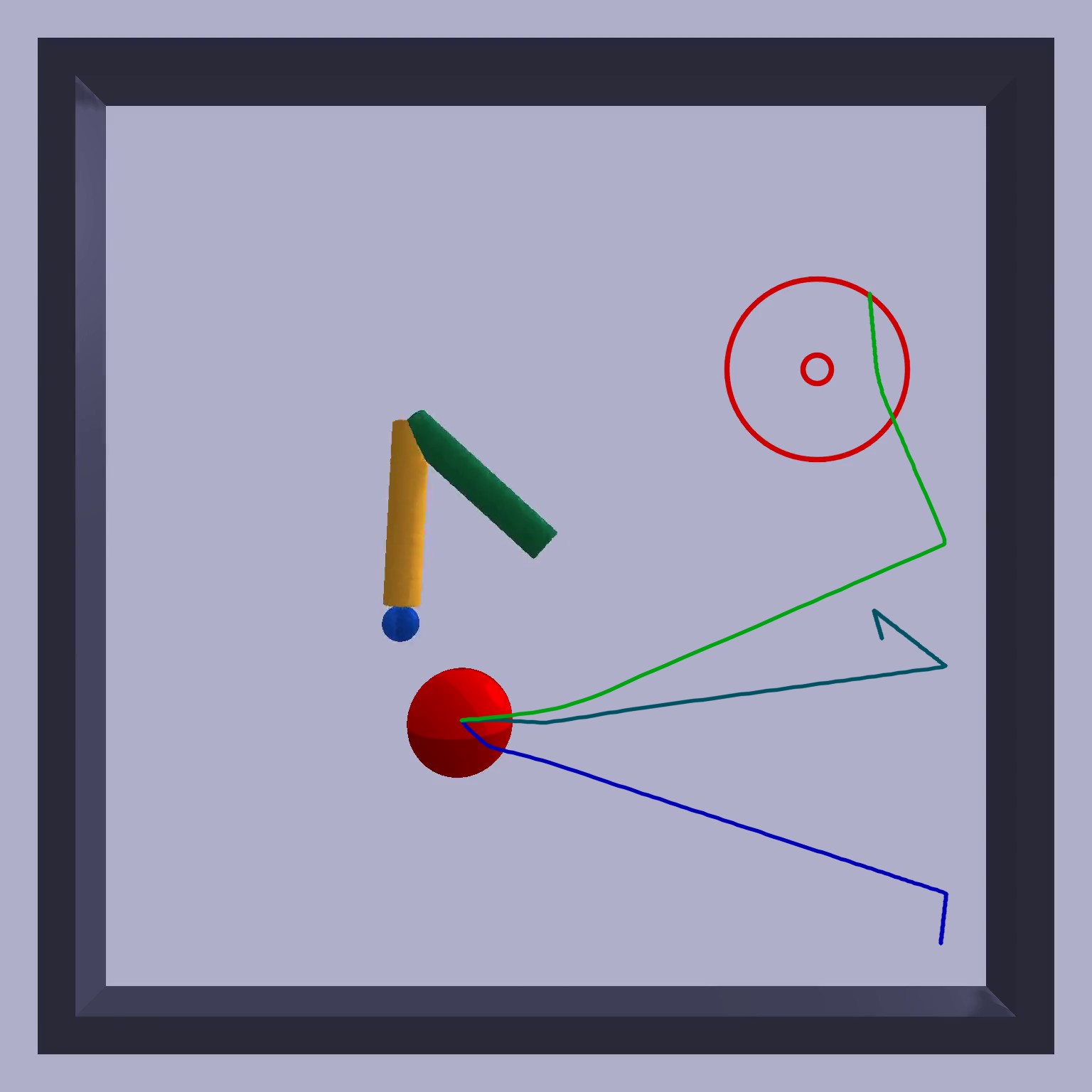}}
\caption{Visualization of the pusher robot environment with illustrated ball trajectories for three different hidden parameter values.}
\label{fig-pusher}
\end{center}
\end{figure}

The goal of the pusher robot environment, shown in Figure \ref{fig-pusher}, is to control a robotic arm to push a ball on to a target location. It is a modified version of the reacher environment from \citet{brockman2016gym}. The state space consists of 12 variables containing positions, orientations,  velocities, and rotations of the arm components and ball. The action space consists of two torque values applied to the two arm joints (shoulder and elbow). The hidden parameter in this environment is the relative strength of the joint torque and can take on a value of 50\%, 75\%, and 100\% of the original strength. Figure \ref{fig-pusher} shows the difference in ball locations of trajectories produced with the same set of actions for different strength values.

\subsection{UAV ISR}
The goal of the UAV ISR environment \cite{conlon2022generalizing} is to control a UAV to fly to a target location in order to collect data. We modify the environment to include a hidden parameter in two ways. One way is by introducing a payload weight sampled from a continuous uniform distribution. The other is the ambient temperature sampled from a continuous distribution. In this work, we assume there is only a single hidden parameter, so when analyzing one hidden parameter, we allow the other to be visible by making it a part of the agent state.

\subsection{Implementation Details}
We use a similar world model architecture (shown in Figure \ref{fig-world-model}) for both of our reinforcement learning environments. It is comprised of a gated recurrent unit (GRU) \citep{chung2014empirical}, which introduces recurrence to the model, as well as a pair of multi-layer perceptrons (MLP), which serve as an encoder and decoder of the state. Since our goal is to filter out transient information from the hidden memory, we introduce a skip connection around the GRU to allow information to propagate from the input state and action to the prediction without going through the memory. An alternative to the skip connection is to partition the hidden memory into a time-invariant portion and a regular portion. The time-invariant portion is trained with our algorithm to represent hidden parameters and the regular portion is trained with the standard training algorithm. Not only does this allow the transient information to be passed from state and action to the prediction, it also allows the GRU to capture information related to non-constant latent variables, which exist in many applications. 

\begin{figure}[ht]
\begin{center}
\centerline{\includegraphics[width=0.9\columnwidth]{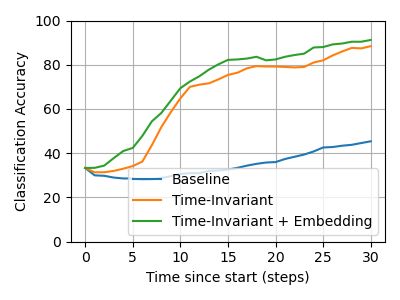}}
\caption{Classification accuracy for mountain car hidden gravity parameter.}
\label{fig-mountain-acc}
\end{center}
\end{figure}

\begin{figure}[ht]
\begin{center}
\centerline{\includegraphics[width=0.9\columnwidth]{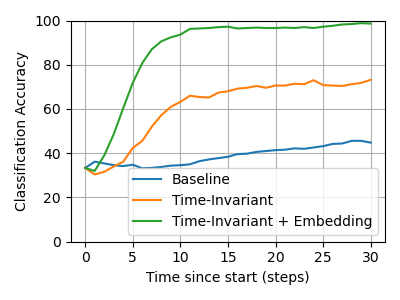}}
\caption{Classification accuracy for pusher hidden strength parameter.}
\label{fig-pusher-acc}
\end{center}
\end{figure}

\begin{figure}[ht]
\begin{center}
\centerline{\includegraphics[width=0.9\columnwidth]{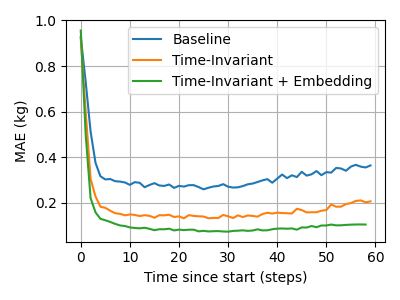}}
\caption{Regression error for hidden UAV payload mass parameter.}
\label{fig-pl-acc}
\end{center}
\end{figure}

\begin{figure}[ht]
\begin{center}
\centerline{\includegraphics[width=0.9\columnwidth]{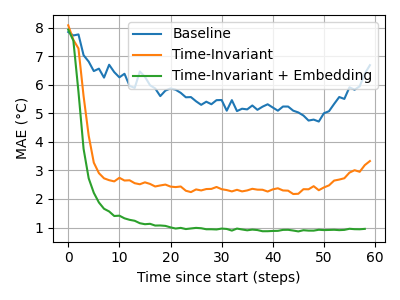}}
\caption{Regression error for hidden UAV ambient temperature parameter.}
\label{fig-temp-acc}
\end{center}
\end{figure}

\subsection{Results}
\subsubsection{Hidden Parameter Estimation}
\label{section-classification}
To demonstrate the effectiveness of our approach we show that our learned features can be used to estimate the hidden parameter of a trajectory with limited labeled training data. For each application, we generate 5000 trajectories and use them to train two world models: one with the baseline RNN training algorithm and a second with our time-invariant memory algorithm. Additionally, we use the same data set to embed the time-invariant memory in a latent bisimulation metric space.

After training the models and learning the embeddings, we collect two smaller data sets: one set of 30 trajectories and one set of 100 trajectories to serve as the training and testing data for the classification task. We use the world models and learned embeddings to generate features for the trajectories and use a k-nearest neighbors to classify the test trajectories. Since our world model produces a new memory output at each time step, we also can generate a new representation of the trajectory at each time step. Therefore, we evaluate the estimation accuracy as a function of the time since the beginning of the trajectory. As one would expect, as the agent interacts with the environment, it acquires more information about the hidden parameters, which increases accuracy.

Our estimation results are shown in Figures \ref{fig-mountain-acc}, \ref{fig-pusher-acc}, \ref{fig-pl-acc}, and \ref{fig-temp-acc} and compare three feature representations. The ``Baseline'' results use the hidden memory of a world model trained with a standard training algorithm, the ``Time-Invariant'' results use the hidden memory of a world model trained with our time-invariant approach, and the ``Time-Invariant + Embedding'' results use the latent bisimulation metric features. It is clear that our approaches significantly outperform the baseline for all environments and parameters, both in terms peak accuracy and the time it takes to reach peak accuracy. The latent bisimulation metric features provide a marginal improvement over the time-invariant features on mountain car, while they offer a significant performance boost on pusher and UAV. We believe this is due to the relative complexity of the environments. Compared to the mountain car environment, the pusher and UAV have much larger states (12+ vs 2 variables), so the world models learn exponentially more patterns to recognize the effects of the hidden parameter. This results in a more complicated hidden memory feature space where values are not easily directly comparable, which benefits more from the metric learning approach. Mountain car is simpler and so the metric learning approach does not add as much value.

\begin{figure*}[ht]
\begin{center}
\centerline{\includegraphics[trim={0cm, 0cm, 0cm, 0cm}, clip, width=0.8\textwidth]{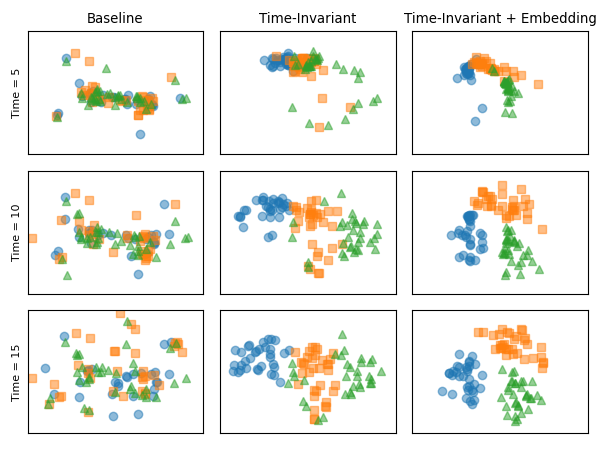}}
\caption{Visualization of latent feature spaces for pusher showing the clean separation of the three different hidden parameter values.}
\label{fig-pusher-tsne}
\end{center}
\end{figure*}

\begin{figure*}[ht]
\begin{center}
\centerline{\includegraphics[trim={0cm, 0cm, 0cm, 0cm}, clip, width=0.8\textwidth]{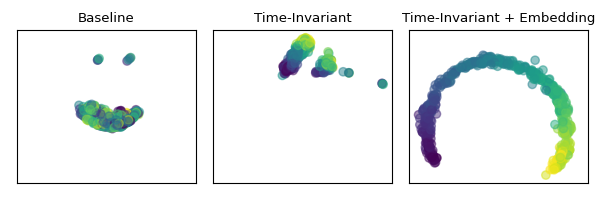}}
\caption{Visualization of latent feature spaces for (continuous) UAV payload mass.}
\label{fig-pl}
\end{center}
\end{figure*}

\begin{figure*}[ht]
\begin{center}
\centerline{\includegraphics[trim={0cm, 0cm, 0cm, 0cm}, clip, width=0.8\textwidth]{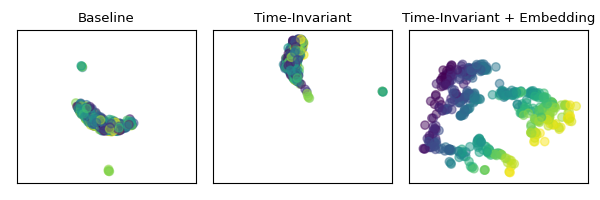}}
\caption{Visualization of latent feature spaces for (continuous) UAV ambient temperature.}
\label{fig-temp}
\end{center}
\end{figure*}

\subsubsection{Latent Space Visualization}
Another way to see the effectiveness of our approach is to visualize the feature spaces learned by the algorithms for the pusher and UAV environments. Since they are too high-dimensional to visualize directly, we further reduce their dimension using unsupervised techniques (t-SNE \citep{van2008visualizing} for pusher and UMAP \citep{mcinnes2018umap} for UAV). 

Figure \ref{fig-pusher-tsne} shows the visualization of our learned features for the pusher environment. We show the features at three times along a trajectory: after 5, 10, and 15 steps. As with the classification results, the ``Baseline'' method uses the hidden memory of a world model trained with a standard training algorithm, the ``Time-Invariant'' method uses the hidden memory of a world model trained with our time-invariant approach, and the ``Time-Invariant + Embedding'' method uses the latent bisimulation metric features. Each plotted marker represents a trajectory at that time, where the marker used signifies the value of the hidden parameter: blue circles, orange squares, and green triangles represent torque strengths of 50\%, 75\%, and 100\% respectively.

Ideally, all trajectories with the same hidden parameter value would be represented by the same point in the feature space and trajectories with different hidden parameters would be far apart from one another. At all three time steps, the ``Baseline'' features perform the worst as the visualization appears random. The ``Time-Invariant'' features display more structure with some overlap between hidden parameter values. The ``Time-Invariant + Embedding'' features clearly perform the best with the most structure and the least overlap. By time-step 15, the three hidden parameter values appear to be easily separable.

Figures \ref{fig-temp} and \ref{fig-pl} show the visualizations for the UAV hidden parameters---payload mass and ambient temperature---respectively. Both feature spaces are well organized by hidden parameter. It is especially noteworthy that the feature space representing the payload mass hidden parameter forms a one-dimensional manifold, as this is the dimension of the original hidden parameter. 

\subsection{Understanding Hidden Parameters}

In this section, we present two methods to help understand better the effects of hidden parameters on systems. First, we show how to determine which state variables are most effected by the hidden parameter. Then, we show how to use the world model to generate ``imagined'' trajectories for a range of hidden parameter values.

\subsubsection{Hidden Parameter State Effects}

Our goal is to understand what state features are most affected by the hidden parameters. We accomplish this by comparing the performance of a stateful world model (an RNN) with a non-stateful world model (an MLP). While neither world model is given explicit information about the hidden parameter, stateful world models can learn to infer information about it from the system’s behavior and use it to make better predictions. More specifically, we compute the ratio (RNN / MLP) of the prediction accuracies for the two models for each feature in the agent’s state to determine which are most affected by the hidden parameters. We then conclude that the hidden parameters have the greatest effect on features for which a stateful world model improves predictions the most and have the least/no effect on features for which the performance is unchanged. 

Tables \ref{tab:pl} and \ref{tab:temp} show the relative prediction accuracies of the two models on each feature in the state. For the hidden payload mass, it is clear that the hidden parameter has the biggest effect on the velocity in the z direction as well as the pitch. This makes sense as the increased gravitational force should have a significant effect on those features. Additionally, the velocity in the x and y direction, the angular roll and pitch velocities, and the position in the z direction are all affected. For the hidden ambient temperature, the battery level is most affected. This makes sense as the battery life is very sensitive to the ambient temperature.

\begin{table}[ht]
\begin{center}
\begin{tabular}{|*{2}{c|}}
    \hline
    Feature & Error Ratio\\\hline
    Position x & \gradient{2.42}\\\hline
    Position y & \gradient{2.36}\\\hline
    Position z & \gradient{4.82}\\\hline
    Roll & \gradient{1.72}\\\hline
    Pitch & \gradient{2.16}\\\hline
    Yaw & \gradient{1.78}\\\hline
    Velocity x & \gradient{2.46}\\\hline
    Velocity y & \gradient{2.39}\\\hline
    Velocity z & \gradient{5.25}\\\hline
    Velocity Roll & \gradient{1.71}\\\hline
    Velocity Pitch & \gradient{1.71}\\\hline
    Velocity Yaw & \gradient{1.35}\\\hline
    Battery & \gradient{0.98}\\\hline
\end{tabular}
\end{center}
\caption{\label{tab:pl}Effect of payload mass on UAV state variables.}
\end{table}

\begin{table}[ht]
\begin{center}
\begin{tabular}{|*{2}{c|}}
    \hline
    Feature & Error Ratio\\\hline
    Position x & \gradient{1.52}\\\hline
    Position y & \gradient{1.51}\\\hline
    Position z & \gradient{2.71}\\\hline
    Roll & \gradient{1.17}\\\hline
    Pitch & \gradient{1.20}\\\hline
    Yaw & \gradient{1.27}\\\hline
    Velocity x & \gradient{1.59}\\\hline
    Velocity y & \gradient{1.60}\\\hline
    Velocity z & \gradient{2.70}\\\hline
    Velocity Roll & \gradient{1.02}\\\hline
    Velocity Pitch & \gradient{1.00}\\\hline
    Velocity Yaw & \gradient{1.06}\\\hline
    Battery & \gradient{8.26}\\\hline
\end{tabular}
\end{center}
\caption{\label{tab:temp}Effect of ambient temperature on UAV state variables.}
\end{table}

\subsubsection{Visualizing Trajectories by Hidden Parameter}

\begin{figure*}[ht]
\begin{center}
\centerline{\includegraphics[trim={0cm, 0cm, 0cm, 0cm}, clip, width=0.8\textwidth]{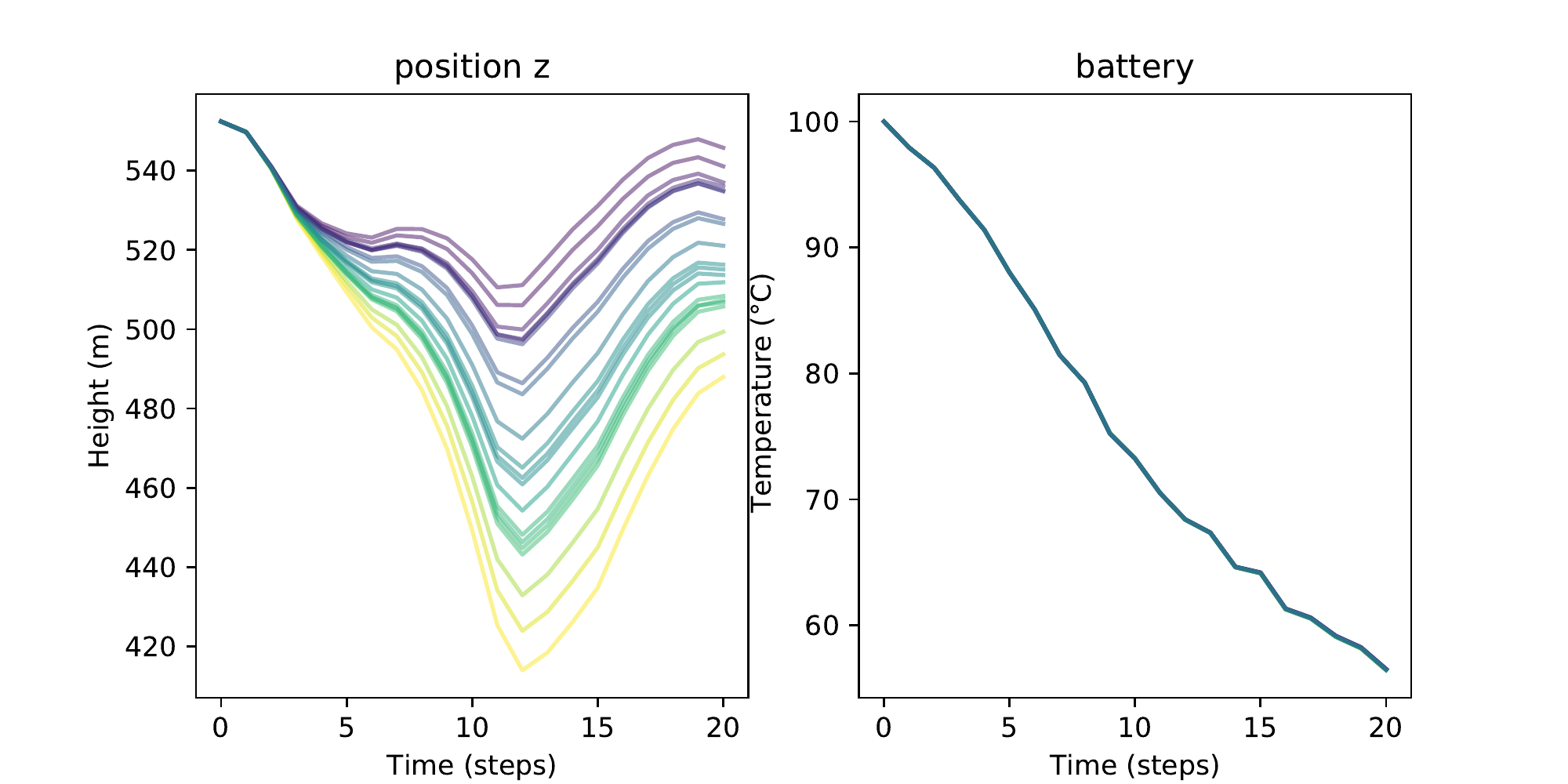}}
\caption{Visualized imagined trajectories for varying UAV payload mass.}
\label{fig-imagine-pl}
\end{center}
\end{figure*}

\begin{figure*}[ht]
\begin{center}
\centerline{\includegraphics[trim={0cm, 0cm, 0cm, 0cm}, clip, width=0.8\textwidth]{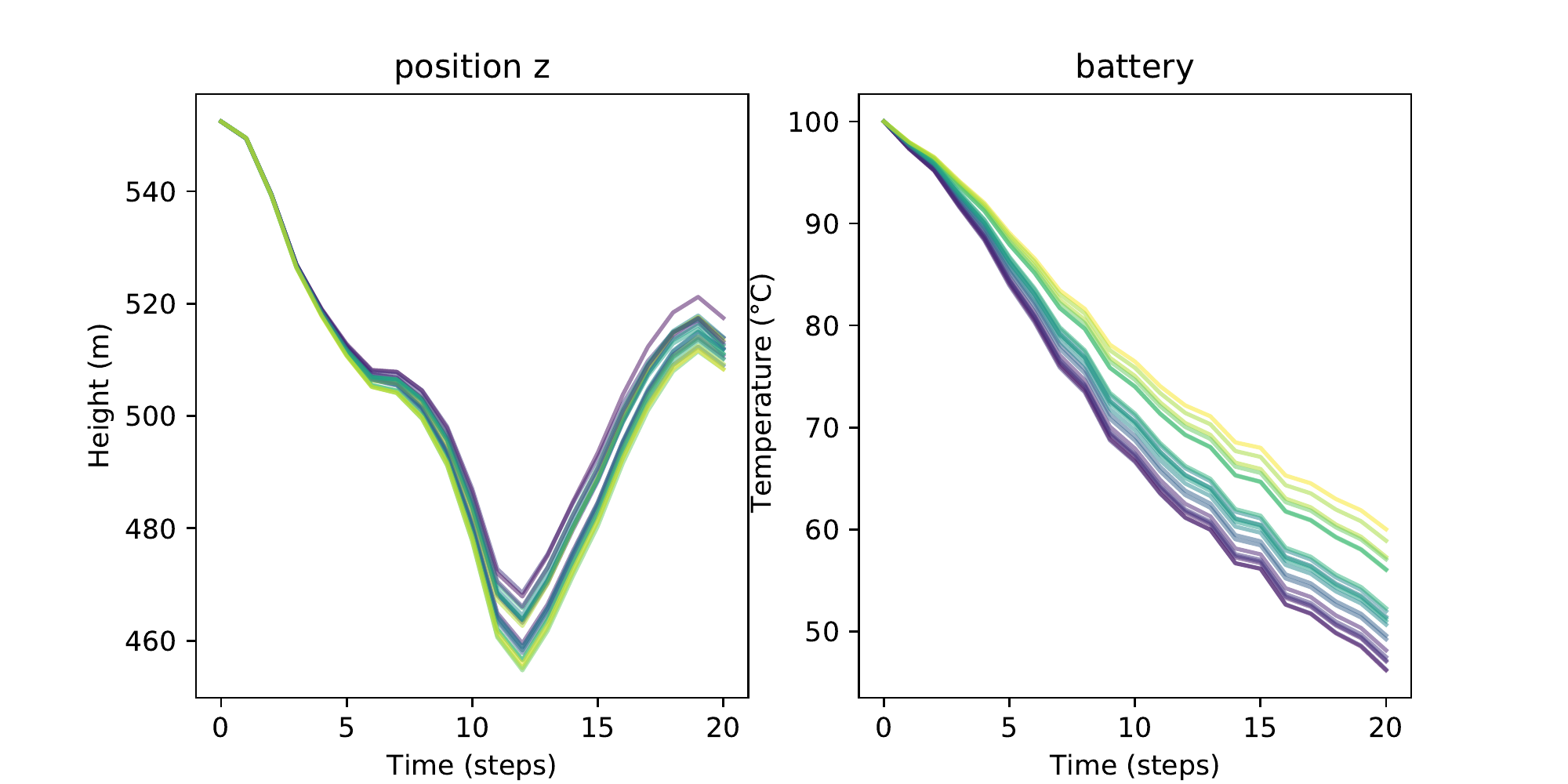}}
\caption{Visualized imagined trajectories for varying UAV ambient temperature.}
\label{fig-imagine-temp}
\end{center}
\end{figure*}

In this section, we show how to visualize ``imagined'' trajectories to enable a user to recognize patterns. Specifically, for a given initial state and sequence of actions, we use the world model roll out trajectories for a variety of hidden parameter values. We then plot the trajectories on the same plots as shown in Figures \ref{fig-imagine-pl} and \ref{fig-imagine-temp}. Note the color of each trajectory indicates the corresponding color in Figures \ref{fig-pl} and \ref{fig-temp}. Figure \ref{fig-pl} shows the trajectories when the payload mass is the hidden parameter. As one would expect, different payload masses effect the position and velocity of the UAV. Figure \ref{fig-temp} shows the trajectories when the temperature is the hidden parameter. The plots show that temperature has a large effect on the battery life but otherwise does not affect the dynamics much.

\section{Conclusions and Future Research}

We have proposed an unsupervised learning approach to represent trajectories by features signifying the hidden parameter of the environment. Our approach consists of two steps. First, we use a novel world model training algorithm to isolate hidden parameter information in an RNN-based world model memory by enforcing a time-invariance constraint. We then use a metric learning approach to map the the RNN-hidden memory to a latent bisimulation metric space where the distance between points signifies the relative difference in system dynamics due to the hidden parameter. Finally, we presented hidden parameter estimation and visualization results to validate our approach on four hidden parameters across three applications.

While our approach works well on the presented environments, they represent a only a small subset of the challenges that could be faced in real world, and there are many potential avenues for future research. One could learn how to estimate the number of hidden parameters in an environment and ways to decouple them to understand their individual effects. Additionally, one could extend from static hidden parameters to dynamic latent variables that change at different time-scales (e.g., temperature changes on the order of hours, while the wind changes on the order of seconds). And finally, the end goal of this problem is to better characterize the behavior of systems to inform human users. Finding more effective ways to tether information about hidden parameters to real-world phenomena in order to better communicate it remains an open problem.

\section{Acknowledgements}
This material is based upon work supported by the Defense Advanced Research Projects Agency (DARPA) under Contract No. HR001120C0032.  Any opinions, findings and conclusions or recommendations expressed in this material are those of the author(s) and do not necessarily reflect the views of DARPA.

\bibliography{llaama_paper}

\end{document}